\pdfoutput=1

\documentclass[11pt]{article}

\usepackage[final]{acl}

\usepackage{times}
\usepackage{float}
\usepackage{latexsym}
\usepackage{multirow} 
\usepackage{eqnarray,amsmath}
\usepackage{listings}
\usepackage[T1]{fontenc}

\usepackage[utf8]{inputenc}

\usepackage{microtype}

\usepackage{inconsolata}

\usepackage{graphicx}
\usepackage{xcolor}
\usepackage{booktabs}

\title{
Leveraging What's Overfixed: \\Post-Correction via LLM Grammatical Error Overcorrection}

\author{
  Taehee Park\thanks{Equal contribution.}$^{1}$ \quad
  Heejin Do\footnotemark[1]$^{2}$ \quad
  Gary Geunbae Lee$^{1,3}$ \\
  $^{1}$ Graduate School of Artificial Intelligence, POSTECH, South Korea \\
  $^{2}$ ETH Zurich, ETH AI Center \\
  $^{3}$ Department of Computer Science and Engineering, POSTECH, South Korea \\
  \texttt{\{taehpark, gblee\}@postech.ac.kr} \quad
  \texttt{heejin.do@ai.ethz.ch}
}

\begin{document}
\maketitle

\begin{abstract}
Robust supervised fine-tuned small Language Models (sLMs) often show high reliability but tend to undercorrect. They achieve high precision at the cost of low recall. Conversely, Large Language Models (LLMs) often show the opposite tendency, making excessive overcorrection, leading to low precision. To effectively harness the strengths of LLMs to address the recall challenges in sLMs, we propose Post-Correction via Overcorrection (PoCO), a novel approach that strategically balances recall and precision. PoCO first intentionally triggers overcorrection via LLM to maximize recall by allowing comprehensive revisions, then applies a targeted post-correction step via fine-tuning smaller models to identify and refine erroneous outputs. We aim to harmonize both aspects by leveraging the generative power of LLMs while preserving the reliability of smaller supervised models. Our extensive experiments demonstrate that PoCO effectively balances GEC performance by increasing recall with competitive precision, ultimately improving the overall quality of grammatical error correction.


\end{abstract}


\section{Introduction}

\begin{figure}
\centering
\includegraphics[width=\linewidth]{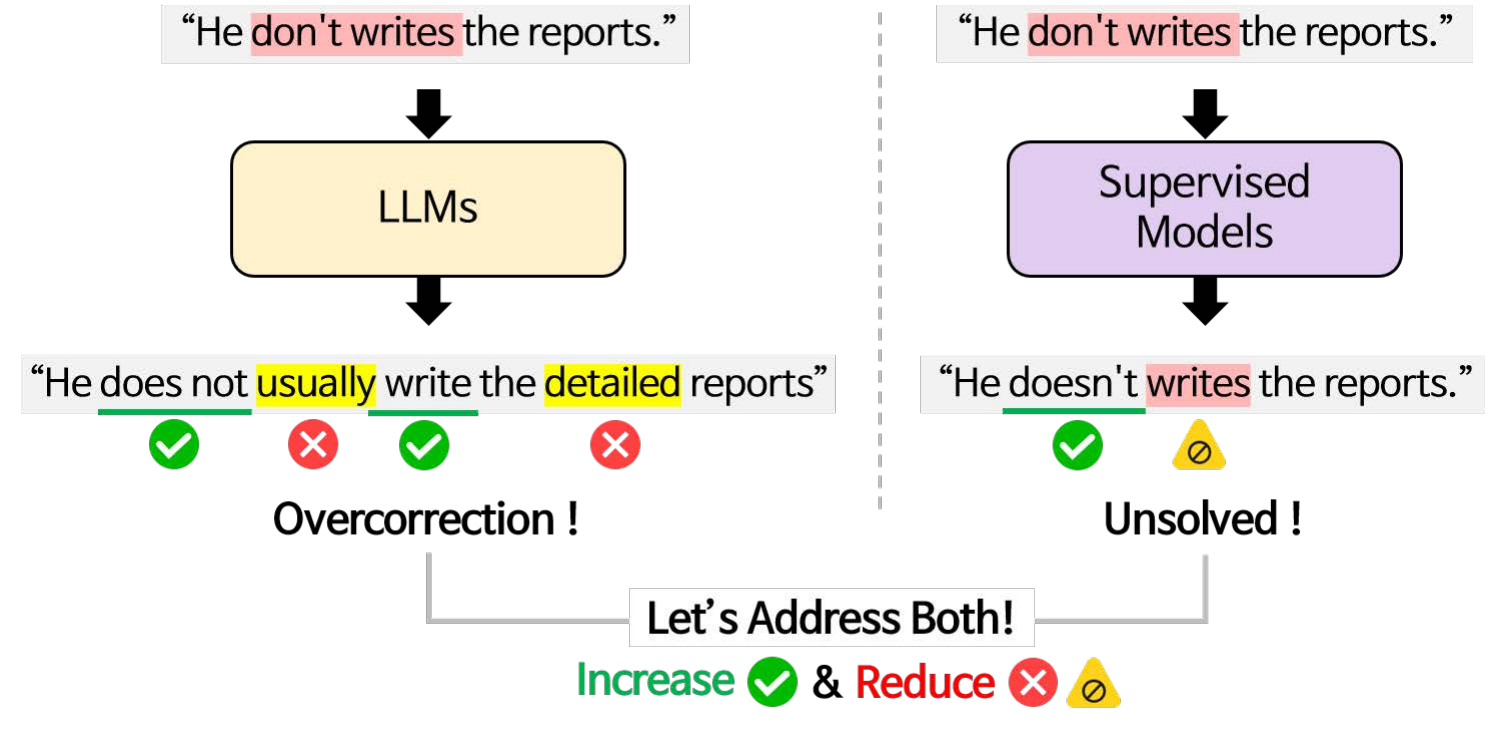}
\caption{Our main motivation: LLMs exhibit high recall but low precision due to overcorrection, while smaller supervised models show high precision but low recall, possibly missing errors. PoCO leverages the strengths of LLMs to address the weaknesses of sLMs.}%
\label{fig: motiv}
\end{figure}

Grammatical error correction (GEC) has become a pivotal task in natural language processing, particularly for language learning applications and writing assistance tools \cite{tetreault-leacock-2014-automated, calo-etal-2021-gecko, kaneko-etal-2022-interpretability, katinskaia-yangarber-2023-grammatical, liang-etal-2023-chatback, 10.4018/IJCALLT.364847}. 
Traditional yet robust fine-tuned small Language Models (sLMs), which have been primarily used for GEC, tend to undercorrect, 
 achieving high precision at the cost of lower recall (Figure~\ref{fig: motiv}). 
 Improving recall is essential in educational settings, as it supports the primary goal of corrective feedback by raising students' awareness of their mistakes \citep{educational_paper}. 
 
 However, existing methods to reduce overcorrection often result in significantly decreased recall, limiting the usability of correction systems \citep{wang-etal-2024-lm}.
This challenge is particularly evident with sLMs, which often exhibit conservative correction tendencies, e.g., making minimal changes or missing errors that require more substantial correction \cite{awasthi-etal-2019-parallel, stahlberg-kumar-2020-seq2edits, katsumata-komachi-2020-stronger, omelianchuk-etal-2020-gector, rothe-etal-2021-simple, tarnavskyi-etal-2022-ensembling}. 
In contrast, Large Language Models (LLMs) exhibit the opposite tendency, e.g., making excessive modifications to input text \cite{loem2023exploringeffectivenessgpt3grammatical, fang2023chatgpthighlyfluentgrammatical, wu2023chatgptgrammarlyevaluatingchatgpt, coyne2023analyzingperformancegpt35gpt4}.
Although LLMs have garnered notable attention for their potential in GEC, owing to their ability to understand and generate human-like text \cite{katinskaia-yangarber-2024-gpt, zeng-etal-2024-evaluating}, their overcorrection often leads to low precision, distorting the intended meaning of the original text and reducing the model's reliability. 



To fully exploit the strengths of LLMs while overcoming the recall challenges in sLMs, we propose Post-Correction via Overcorrection (PoCO), a novel approach that enables effective utilization of LLM outputs within sLM fine-tuning. 
PoCO is a two-step method that first intentionally triggers overcorrection to maximize the possible correction via LLM and then applies a post-correction step to refine and repair the excessive changes. By leveraging the enriched generative power of LLMs while preserving the reliability of sLMs, PoCO harmonizes recall and precision. Thereby, we aim to enhance GEC quality by ensuring comprehensive error revision and high-quality post-corrections.

We conduct experiments to compare PoCO with robust supervised sLMs as well as few-shot, zero-shot, and fine-tuned LLMs.
Our results show that PoCO successfully achieves superior recall scores across all evaluations and maintains competitive precision on both BEA19 development and test sets compared to robust supervised sLMs, highlighting the efficacy of our balancing strategy. 
Notably, when compared with LLMs, PoCO exhibits competitive performance and even higher $F_{0.5}$ scores in the BEA 19 test set.
%

To further validate our approach, we compare PoCO with quality estimation-based filtering techniques like GRECO \cite{qorib-ng-2023-system}, as well as robust GEC ensemble systems, such as ESC \cite{qorib-etal-2022-frustratingly}. The results show that our method outperforms GRECO in a single-model setting and also demonstrates superior performance when integrated into ensemble systems.
Our novel contributions are as follows:




\begin{itemize} 
    \item We address low-recall issues in robust supervised fine-tuned sLMs with PoCO by leveraging the LLMs' overcorrection tendency, a marked limitation of LLM-based GEC. 
    Consequently, PoCO achieves the highest recall among robust fine-tuned single models.
    
    \item To mitigate the typically low precision challenges inherent in LLMs, which reduce the reliability of model outputs, PoCO integrates a novel training strategy centered on a \textit{recovered target}. By incorporating the recovered target, PoCO consistently shows improved precision across all datasets, indicating the efficacy of directly recovering erroneous corrections.
    
    
    
    \item We investigate the effectiveness of PoCO within a quality estimation system and a regression-based ensemble system, where it demonstrates high-performing results. In ESC settings, the experimental evidence indicates that our model can perform better using fewer models, further highlighting its efficiency.
\end{itemize}
\section{Related Work}

\paragraph{GEC}

Single-model frameworks for GEC primarily follow two mainstream approaches: edit-based and Seq2Seq methods. The edit-based approach predicts edit operations \cite{malmi-etal-2019-encode, awasthi-etal-2019-parallel, stahlberg-kumar-2020-seq2edits, omelianchuk-etal-2020-gector, tarnavskyi-etal-2022-ensembling}, while the Seq2Seq method, inspired by machine translation, treats GEC as a monolingual translation task
where erroneous text is rewritten into corrected text \cite{xie2016neurallanguagecorrectioncharacterbased, yuan-briscoe-2016-grammatical, ji-etal-2017-nested, junczys-dowmunt-etal-2018-approaching, Chollampatt_Ng_2018}. Recent advancements, such as \citet{rothe-etal-2021-simple}, have demonstrated the effectiveness of Seq2Seq models like T5 \cite{2020t5}, which generate error-free sentences directly from erroneous inputs, achieving strong performance.

Various other efforts have been made to enhance GEC by incorporating additional components into a strong single-model correction system \cite{yuan-etal-2021-multi, yasunaga-etal-2021-lm, sorokin-2022-improved, zhou-etal-2023-improving-seq2seq}. Recently, ensemble methods that combine multiple models have become increasingly prevalent to further enhance performance \cite{tarnavskyi-etal-2022-ensembling, omelianchuk-etal-2024-pillars, qorib-etal-2024-efficient}. 
For instance, \citet{qorib-ng-2023-system} proposed GRECO, a quality estimation system that scores edits using word label and gap label, while 
ESC \cite{qorib-etal-2022-frustratingly} combines robust GEC models using logistic regression to score each edit. 
However, despite various attempts,
these approaches reveal a clear limitation of low recall. While ensembling enhances precision—a key advantage of supervised fine-tuning—it still fails to address the issue of low recall. This suggests that simply ensembling LMs is insufficient to overcome this fundamental limitation.

\begin{figure*}
\centering
\includegraphics[width=\linewidth]{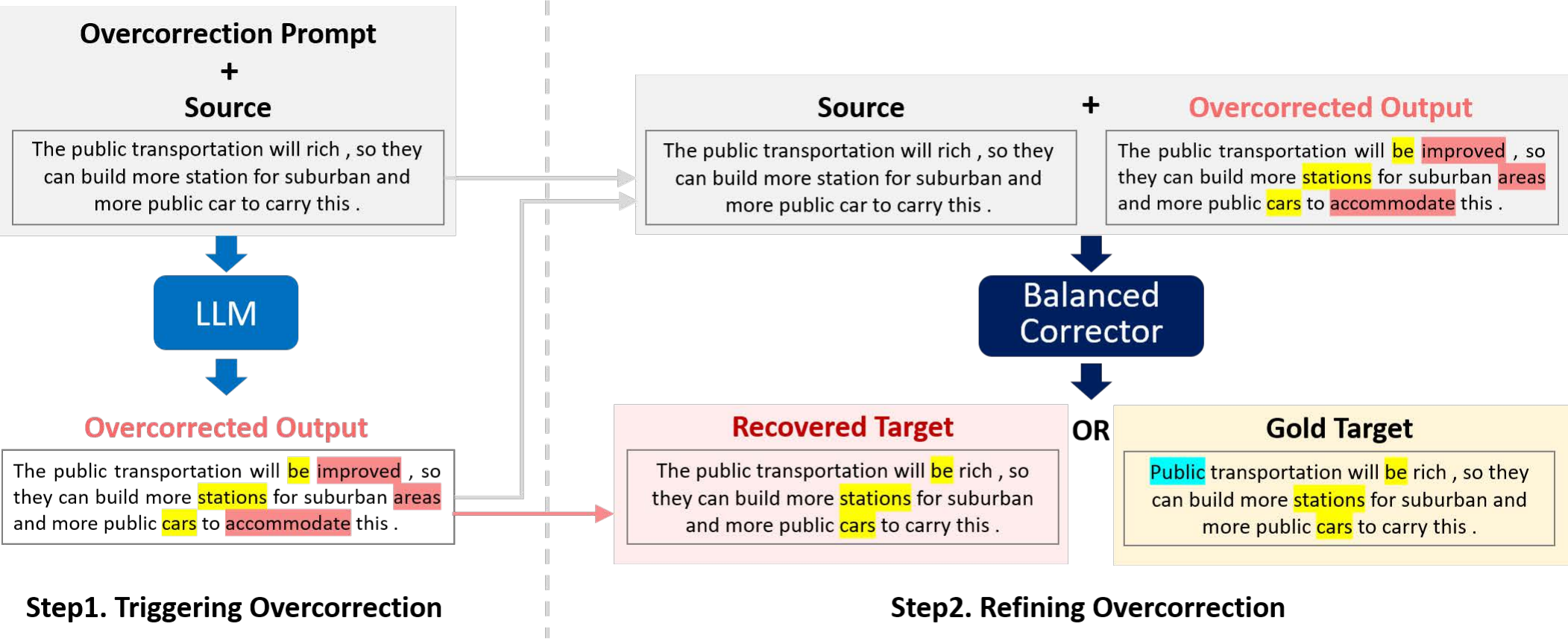}
\caption{Overview of the POCO Framework: The yellow highlight shows the parts of the LLM-overcorrected output that match the gold target, while the pink highlight indicates the parts not in the gold target. The blue color marks the parts of the gold target missing from the LLM output. The "Recovered Target" includes only the yellow-highlighted portions (matching the gold target), with the rest restored to the original source text.}%
\label{fig: main}
\end{figure*}

\paragraph{LLM for GEC}

While Large Language Models (LLMs) have shown remarkable success across various domains, their application to GEC has faced challenges, particularly in precision. Prompting-based approaches using models like GPT and LLaMA \cite{coyne2023analyzingperformancegpt35gpt4, wu2023chatgptgrammarlyevaluatingchatgpt, katinskaia-yangarber-2024-gpt, yang-quan-2024-alirector} are revealed to struggle with overcorrection, leading to unreliable edits. Studies have explored optimizing prompts \cite{loem2023exploringeffectivenessgpt3grammatical} and employing techniques like Chain-of-Thought (CoT) \cite{fang2023chatgpthighlyfluentgrammatical}, but precision remains a key limitation. To mitigate this, recent high-performing approaches fine-tune LLMs such as LLaMA-7B, LLaMA-13B, T5-11B, and UL2-20B \cite{omelianchuk-etal-2024-pillars}, improving overall GEC performance. However, even fine-tuned LLMs tend to exhibit lower precision compared to sLMs, having tendency toward overcorrection.

We propose a novel method that addresses the low recall of sLMs by leveraging complementary strengths of LLMs, which effectively mitigates the limitations of existing methods.

\section{PoCO}
PoCO leverages the overcorrection tendencies of LLMs to address the low recall commonly observed in robust supervised sLMs, while simultaneously preserving their precision through a targeted post-correction step.
Our approach consists of two stages: 1) \textit{Triggering Overcorrection}, which prompts LLM to make extensive corrections, and 2) \textit{Post Correction}, which then refines the overcorrections for improving the overall score.
The overall structure of our method is presented in Figure \ref{fig: main}.

\subsection{Triggering Overcorrection}

At this stage, our primary goal is to \textit{improve recall} by leveraging LLMs' tendency to generate extensive corrections. While inducing overcorrection may temporarily reduce precision and impact overall performance, we address this issue in the next step. Here, our focus is on mitigating the conservative correction (low recall) of the supervised fine-tuned model by intentionally triggering overcorrection.

To achieve high recall, we incorporate the instruction \texttt{``find as many errors as you can''} to maximize the model's correction capability and actively induce overcorrection. This intentional strategy helps capture a broader range of potential errors, which can be refined in later stages. Unlike prior approaches that passively suffer from overcorrection as an unintended side effect, we systematically integrate it as a deliberate step in our pipeline by explicitly encouraging overcorrection for the GEC task. 

Our prompting approach builds on the method proposed by \citet{fang2023chatgpthighlyfluentgrammatical}, which utilizes a detailed prompt that suppresses the model's ability to modify with zero-shot CoT \cite{NEURIPS2022_8bb0d291} reasoning to improve overall performance.
However, since restricting the model’s correction ability contradicts our objective, we removed the phrase \texttt{``keeping the original sentence structure unchanged as much as possible''} from their prompt. 
This modification allows the model to make more flexible corrections, effectively assisting in overcoming the limitations of overly conservative correction approaches in the next step.
Table \ref{tab:prompting} provides the details of our prompt.

\definecolor{yel}{RGB}{255, 240, 153}

\begin{table}[t]
    \centering
    \scalebox{0.7}{
    \begin{tabular}{ p{10cm} }
        \toprule
        \textbf{CoT Prompt }\\
        \hline
        ``Please identify and correct any grammatical errors in the following sentence indicated by \texttt{<input> ERROR </input>} tag, you need to comprehend the sentence as a whole before identifying and correcting any errors step by step \colorbox{pink}{while keeping the original sentence structure unchanged as much} \colorbox{pink}{as possible.} Afterward, output the corrected version directly without any explanations. Remember to format your corrected output results with the tag \texttt{<output> Your Corrected Version </output>}. Please start: \texttt{<input> INPUT </input>:}" \\
        \midrule
        \textbf{Overcorrection CoT Prompt} \\
        \hline
        ``Please identify and correct any grammatical errors in the following sentence, indicated by \texttt{<input> ERROR </input>} tag. You need to comprehend the sentence as a whole and \colorbox{yel}{find as many errors as you can before identifying and correcting} \colorbox{yel}{them step by step}. Afterward, output the corrected version directly without any explanations. Remember to format your corrected output results with the tag \texttt{<output> Your Corrected Version </output>}. Please start: \texttt{<input> INPUT </input>:}" \\
        \bottomrule
    \end{tabular}
    }
    \caption{The original CoT prompt \cite{fang2023chatgpthighlyfluentgrammatical} and our overcorrection triggering prompt used.}
    \label{tab:prompting}
\end{table}

\subsection{Post Correction}
The generated overcorrected outputs would possibly exhibit high recall but low precision. Therefore, to directly enhance the precision, we suggest a double-target training strategy. 
Our approach utilizes two types of targets: (1) the ground truth (gold) target, which is manually annotated by human annotators, and (2) the \textit{recovered target}, which is generated by our novel method.

Each target serves a distinct purpose. Training with the gold target encourages the model to simultaneously learn two tasks: 
The first objective is to restore overcorrected words, and the second is to correct new errors that the LLM failed to address in the first stage.
Providing a gold target can also effectively improve precision while preserving a substantial portion of recall, making it a practical approach to mitigating overcorrection.
However, since precision is often prioritized over recall in the GEC task, we introduce an additional technique, \textit{recovered target}, to further enhance precision, which was intentionally reduced in Section 3.1.

\textit{Recovered target} prioritizes restoring overcorrected parts during training, rather than primarily focusing on correcting new errors. It is constructed by selectively incorporating correct edits from the gold target, but only within the regions modified by the LLM. To guide the model toward more precise corrections, we create two training pairs for each source sentence: one with the gold target and one with the recovered target. 
This simple yet effective integration improves model precision by guiding it to reverse overcorrections while preserving valid edits. By jointly training the model on both the gold and recovered targets, PoCO achieves this balance without significantly compromising recall and precision.
When training the model, we pair the original source sentences with the LLM-generated sentences using the following format, which serves as the input to the encoder-decoder model:

\begin{verbatim}
source : source sentence
overcorrect : LLM-generated sentence
\end{verbatim} 
Our proposed recovered target with the gold target during training could achieve two key objectives: 
(1) Enabling the model to correct errors that the LLM failed to fix using the gold target (2) significantly improving precision, which drops substantially in Step 1 (Triggering Overcorrection), by guiding the model to concentrate more on recovering overcorrected parts.
In addition, we suggest two training strategies. In the PoCO-\textit{Seq} setting, the model is first pre-trained on the gold target and then sequentially fine-tuned on the recovered target. In the PoCO-\textit{Mix} setting, training is conducted using a mixture of gold and recovered target sentences.

\section{Experiment}

\subsection{Datasets}
We use the Clang-8 \cite{rothe-etal-2021-simple}\footnote{\url{https://github.com/google-research-datasets/clang8?tab=readme-ov-file}} and W\&I + LOCNESS \cite{bryant-etal-2019-bea}\footnote{\url{https://www.cl.cam.ac.uk/research/nl/bea2019st/data/wi+locness_v2.1.bea19.tar.gz}} datasets for model training. Clang-8 is a refined version of the Lang-8 Corpus of Leaner English (Lang-8) \cite{tajiri-etal-2012-tense}\footnote{\url{https://sites.google.com/site/naistlang8corpora}} dataset. 
We select the Clang-8 dataset for pretraining following the approach used in previous sequence-to-sequence models \cite{rothe-etal-2021-simple}.
Note that in pretraining, we only use the Clang-8 dataset without using LLM overcorrection results.
Then, to fine-tune our model with LLM-generated overcorrected outputs, we used only the W\&I + LOCNESS dataset; thus, triggering the overcorrection step is also conducted solely on this dataset.

For evaluation, we tested our model on the BEA-19 \cite{bryant-etal-2019-bea} dev and test datasets and the CoNLL-14 \cite{ng-etal-2014-conll} test set. 
The precison, recall, $F_{0.5}$ scores are computed using Errant scorer \cite{bryant-etal-2017-automatic} and the M2 scorer \cite{dahlmeier-ng-2012-better}, respectively. The BEA-19 test set can be evaluated on the BEA-19 Shared Task platform\footnote{\url{https://www.cl.cam.ac.uk/research/nl/bea2019st}}.

\subsection{Settings}
\paragraph{LLMs}
We attempted to replicate the results of previous studies by using GPT-3.5-turbo-0613, the model that shows strong performance in grammatical error correction. However, since OpenAI\footnote{\url{https://platform.openai.com/docs/models}} officially ended support for this version, the model used in the previous study, GPT-3.5-Turbo-0613, is no longer accessible. 
Thus, we conducted our experiments using one of the available versions of the GPT-3.5 series,
GPT-3.5-Turbo-0125 (\texttt{GPT3.5-CoT-0125}). 
To ensure a fair evaluation, we compare the implementation with and without our overcorrection strategy, using the same version of the model to validate the effectiveness of our approach.
%
In previous studies, the GPT-3.5 model generally outperformed the GPT-4 model in terms of recall. Therefore, in the process of inducing overcorrection, our experiments were conducted exclusively using the GPT-3.5 model. 
The temperature is set to 1.

%

\paragraph{Fine-tuning}
Our model is primarily based on the encoder-decoder architecture.
In particular, we conducted all the experiments based on the open-source T5 model\footnote{\url{https://github.com/gotutiyan/gec-t5}}, which had been pretrained on the Clang-8 dataset \cite{rothe-etal-2021-simple}. The target data is described in Section 3.2. The model is fine-tuned with a learning rate of 1e-4, batch size of 64, and 10 epochs.



\subsubsection{Baseline models}

\paragraph{Fine-tuned sLMs}
For robust fine-tuning models, we include not only single models such as Seq2Edits \cite{stahlberg-kumar-2020-seq2edits}, Tagged-Corruption \cite{stahlberg-kumar-2021-synthetic}, and MoECE-GS-Large \cite{qorib-etal-2024-efficient}, but also approaches that enhance performance by attaching auxiliary scorer like EditScorer \cite{sorokin-2022-improved}.
For EditScorer, CTC-Copy, and GECToR-2024, we report the results re-evaluated by \citet{omelianchuk-etal-2024-pillars}. EditScorer, originally proposed by \citet{sorokin-2022-improved}, is evaluated based on its open-source\footnote{\url{https://github.com/AlexeySorokin/EditScorer}} implementation, using the GECoR-XLNet$^{(L)}$ option from \citet{tarnavskyi-etal-2022-ensembling} with a RoBERTa-Large encoder as the scorer.
CTC-Copy is initially introduced by \citet{zhang-etal-2023-non}, and the reported performance in our paper is the implementation of official code\footnote{\url{https://github.com/yzhangcs/ctc-copy}} using a RoBERTa encoder.
GECToR-2024 is the method originally proposed by \citet{omelianchuk-etal-2020-gector}, but reported performance in our result is a new version of GECTor where additional training was applied to GECToR-RoBERTa$^{(L)}$ to further enhance performance.

\paragraph{LLMs}
When comparing our overcorrection method, we got results reported by \citet{omelianchuk-etal-2024-pillars} and \citet{katinskaia-yangarber-2024-gpt}. 
Since \citet{fang2023chatgpthighlyfluentgrammatical} did not explicitly specify the used GPT version, we report results from \citet{katinskaia-yangarber-2024-gpt} using the same prompt with \citet{fang2023chatgpthighlyfluentgrammatical}.
%
In post-correction, we compare our model against two types of LLM baselines: (1) LLMs that enhance performance through few-shot prompting and (2) LLMs fine-tuned specifically for the GEC task. GPT-3.5 few-shot prompting baselines were referenced from \citet{fang2023chatgpthighlyfluentgrammatical}, and the 16-shot GPT-3 setting results were taken from \citet{loem2023exploringeffectivenessgpt3grammatical}. For the fine-tuned LLM baseline, we use the results reported in \citet{omelianchuk-etal-2024-pillars}, where each fine-tuned LLMs were leveraged in an ensemble to achieve state-of-the-art performance.

\begin{table}[t]
    \centering
    \scalebox{0.72}{
    \begin{tabular}{l | c  c c | c c c}
        \toprule
        \multirow{2}{*}{Model} & \multicolumn{3}{c|}{CoNLL-14 test} & \multicolumn{3}{c}{BEA-19 dev} \\ 
         & P & R & $F_{0.5}$ & P & R & $F_{0.5}$ \\ \hline
        Chat-LLama2-7B [1] & 42.9 & 47.3 & 43.7 & 19.1 & 34.1 & 21.0 \\ 
        Chat-LLama2-13B [1]& 49.1 & 56.1 & 50.4 & 30.6 & 45.0 & 32.7 \\ 
        GPT3.5{\small-0613} [1]& 56.2 & 57.7 & 56.5 & 37.4 & 50.6 & 39.4 \\ 
        GPT4 [1]& 59.0 & 55.4 & 58.2 & 42.5 & 45.0 & 43.0 \\
        GPT3.5-CoT{\small-0613} [1] & 56.0 & 58.7 & 56.5 & 36.4 & 50.8 & 38.5 \\ 
         
        GPT3.5-CoT{\small-0613} [2] & 55.8 & 58.5 & 56.3 & - & - & - \\ 
         \midrule
         
        GPT3.5-CoT{\small-0125} [2]* & 59.2 & 53.9 & 58.0 & 52.8 & 51.7 & 52.6 \\ 
        \textbf{GPT3.5-CoT-{O.C}{\small-0125}} & 55.0 & {57.9} & 55.5 & 48.2 & {54.8 }& 49.4 \\ \bottomrule
    \end{tabular} 
    }
    \caption{Performance of LLM Zero-shot on the CoNLL-14 test and BEA-19 dev datasets. [1] indicates results reported by \citet{omelianchuk-etal-2024-pillars}, while [2] refers to results following the prompt proposed by \citet{fang2023chatgpthighlyfluentgrammatical}. 
    * denotes our implementation using the currently available version of GPT (GPT-3.5-0125), and O.C denotes the inclusion of our overcorrection strategy. P and R are precision and recall, respectively.}
    \label{tab:prompting_results}
\end{table}

\section{Result}

\subsection{Triggering Overcorrection}



\begin{table*}[h!]
    \centering
    \scalebox{0.78}{
    \begin{tabular}{lccc|ccc|ccc|c}
        \toprule
        \multirow{2}{*}{\textbf{Model}} & \multicolumn{3}{c}{\textbf{CoNLL-14 Test}} & \multicolumn{3}{c}{\textbf{BEA-19 Dev}} & \multicolumn{3}{c}{\textbf{BEA-19 Test}} & \multirow{2}{*}{\shortstack{\textbf{Training Data} \\ \textbf{Size (\#tokens)}}} \\
        
        & Precision & Recall & F$_{0.5}$ & Precision & Recall & F$_{0.5}$ & Precision & Recall & F$_{0.5}$ & \\
        \midrule
        Seq2Edits &  69.9 &  44.4 & 62.7 & - & - & - & 72.7 & 62.9 & 70.5 & - \\

        Tagged-Corruption &  {72.8} &  49.5 & \underline{66.6} & 59.5 & 41.3 & 54.7 & 72.1 & 64.4 & 70.4 & - \\
        
        EditScorer & \textbf{78.5} & 39.4 & 65.5 & \textbf{67.3} & 36.1 & \underline{57.4} & \textbf{81.0} & 56.1 & 74.4 & 96.57M \\
        CTC-Copy & 72.6 & 47.0 & 65.5 & 58.3 & 38.0 & 52.7 & 71.7 & 59.9 & 69.0 & 30.24M \\
        
        GECToR-2024 & \underline{75.0} & 44.7 & 66.0 & \underline{64.6} & 37.2 & 56.3 & 77.7 & 59.0 & 73.1 & 124.57M \\
        MoECE-GS-Large  & 74.3 &  \underline{50.2} & \textbf{67.8} & - & - & 56.4 & 76.9 &  \underline{64.5} & 74.1 & 28.00M \\
        \midrule
        \textbf{PoCO-\textit{Seq}} & 70.2 & 48.9 & 64.6 & {63.0} & \underline{42.4} & \underline{57.4} & \underline{78.6} & 64.3 & \underline{75.3} & 28.62M \\
        \textbf{PoCO-\textit{Mix}} & 69.5 & \textbf{51.9} & 65.1  & {62.3} & \textbf{45.3} & \textbf{57.9}  & {78.0} & \textbf{67.8} & \textbf{75.7} & 28.62M \\
        \bottomrule
    \end{tabular}
    }
    \caption{Performance comparison between supervised fine-tuned models and PoCO (large) across different test sets.}
    \label{tab:performance4}
\end{table*}

\begin{table*}[ht]
    \centering
    \scalebox{0.78}{
    \begin{tabular}{l|ccc|ccc|cccc}
        \toprule
        \multirow{2}{*}{\textbf{Model}} & \multicolumn{3}{c|}{\textbf{CoNLL-14 Test}} & \multicolumn{3}{c|}{\textbf{BEA-19 Dev}} & \multicolumn{4}{c}{\textbf{BEA-19 Test}} \\ 
        & \textbf{Precision} & \textbf{Recall} & \textbf{F$_{0.5}$} & \textbf{Precision} & \textbf{Recall} & \textbf{F$_{0.5}$} & \textbf{Precision} & \textbf{Recall} & \textbf{F$_{0.5}$} & \textbf{Var} \\ 
        \midrule

        Triggered Overcorrection & 55.1 & 57.9 & 55.6 & 48.3 & 54.8 & 49.4 & 51.9 & 69.4 & 54.6 & -\\ \midrule

        PoCO-base-\textit{Gold} & 66.2 & \textbf{51.4} & \underline{62.6} & 56.9 & \textbf{44.1} & 53.8 & 73.1 & \textbf{67.9} & 72.0  & -\\ 
        PoCO-base-\textit{Recovered} & 69.0 & 44.0 & 62.0 & 61.8 & 36.9 & 54.4 & \underline{78.0} & 59.8 & 73.6 & - \\ 
        \textbf{PoCO-base-\textit{Seq}} & \textbf{69.4} & 44.5 & 62.4 & \textbf{63.0} & 38.2 & \underline{55.8} & \textbf{78.8} & 60.2 & \underline{74.2} & - \\ 
        \textbf{PoCO-base-\textit{Mix}}& \underline{69.2} & \underline{49.1} & \textbf{63.2} & \underline{62.5} & \underline{43.2} & \textbf{57.4} & 77.2 & \underline{65.7} & \textbf{74.6} & 74.47 ± 0.22 \\ \midrule
        PoCO-large-\textit{Gold} & 68.6 & \textbf{55.1} & \textbf{65.4} & 59.0 & \textbf{47.9} & {56.3} & 73.9 & \textbf{69.7} & 73.0 & - \\ 
        PoCO-large-\textit{Recovered} & \underline{69.8} & 48.9 & 64.3 & 62.2 & 41.4 & 56.5 & \underline{78.2} & 63.4 & 74.7 & - \\ 
        \textbf{PoCO-large-\textit{Seq}} & \textbf{70.2} & 48.9 & 64.6 & \textbf{63.0} & 42.4 & \underline{57.4} & \textbf{78.6} & 64.3 & \underline{75.3} & - \\ 
        \textbf{PoCO-large-\textit{Mix}} & 69.5 & \underline{51.9} & \underline{65.1} & \underline{62.3} & \underline{45.3} & \textbf{57.9} & 78.0 & \underline{67.8} & \textbf{75.7} & 75.64 ± 0.1 \\ \bottomrule
    \end{tabular}
    }
    \caption{Results comparing the impact of different target combinations. Triggered overcorrection indicates the results of {GPT3.5-CoT-{O.C}{\small-0125}} in Table~\ref{tab:prompting_results}. Variance values (Var) of the BEA-19 test are computed across three independent runs with different random seeds.}
    \label{tab:results5}
\end{table*}

The results in Table \ref{tab:prompting_results} highlight the effectiveness of our triggering overcorrection prompting method in improving recall while maintaining competitive precision. 
Our method shows consistent improvement in recall scores compared to the prompt proposed in \citet{fang2023chatgpthighlyfluentgrammatical} when applied to GPT-3.5-Turbo-0125 across both the CoNLL-14 test and BEA-dev datasets. 
Furthermore, excluding recall derived from the unavailable GPT-3.5-Turbo-0613, our approach achieves the highest recall in both datasets.

\subsection{PoCO Main Results}
\paragraph{Comparison with sLMs}
We successfully address the low recall issue of robust, smaller, supervised fine-tuning models. As shown in Table \ref{tab:performance4}, our model achieves the highest recall on the CoNLL-14 test set and BEA-19 dev and test sets. Additionally, it also records the highest $F_{0.5}$ scores on both BEA-19 dev and test sets.
Additionally, our approach demonstrates advantages in terms of data efficiency. Our method pretrains the model using the CLang-8 dataset, the dataset used in MoECE-GS-Large \cite{qorib-etal-2024-efficient}, and fine-tunes on a relatively small W \& I dataset with a token size of 628.7k.
Unlike previous studies that attempted to enhance model performance by generating large-scale synthetic data, our approach demonstrates that fine-tuning a robust model with a small dataset can achieve significant performance improvements. 


\begin{table*}[h!]
    \centering
    \scalebox{0.74}{
    \begin{tabular}{lccc|ccc|cccc}
        \toprule
        \multirow{2}{*}{\textbf{Model}} & \multicolumn{3}{c}{\textbf{CoNLL-14 Test}} & \multicolumn{3}{c}{\textbf{BEA-19 Dev} } & \multicolumn{3}{c}{\textbf{BEA-19 Test}} \\
        
        & Precision & Recall & F$_{0.5}$ & Precision & Recall & F$_{0.5}$ & Precision & Recall & F$_{0.5}$ \\
        \midrule
        GPT-3.5-1-shot-CoT \cite{fang2023chatgpthighlyfluentgrammatical} & 52.0 & 58.1 & 53.1 & 42.5 & \textbf{55.6} & 44.6 & 34.6 & 69.7 & 38.4 \\
        GPT-3.5-3-shot-CoT \cite{fang2023chatgpthighlyfluentgrammatical} & 51.3 & \textbf{62.4} & 53.2 & - & - & - & 34.0 & \underline{70.2} & 37.9 \\
        GPT-3.5-5-shot-CoT \cite{fang2023chatgpthighlyfluentgrammatical} & 50.9 & \underline{61.8} & 52.8 & - & - & - & 32.4 & 69.9 & 36.3 \\
        GPT-3-16-shot \cite{loem2023exploringeffectivenessgpt3grammatical}& - & - & 57.1 & - & - & - & - & - & 57.4 \\
        \midrule
        Chat-LLaMa-2-7B-FT \cite{omelianchuk-etal-2024-pillars} & \underline{75.5} & 46.8 & 67.2 & 58.3 & 46.0 & 55.3 & 72.3 & 67.4 & 71.2 \\
        Chat-LLaMa-2-13B-FT \cite{omelianchuk-etal-2024-pillars} & \textbf{77.3} & 45.6 & \textbf{67.9} & 59.8 & 46.1 & 56.4 & 74.6 & 67.8 & 73.1 \\
        T5-11B \cite{omelianchuk-etal-2024-pillars} & {70.9} & 56.5 & \underline{67.5} & 60.9 & \underline{51.1} & \textbf{58.6} & 73.2 & \textbf{71.2} & 72.8 \\
        UL2-20B \cite{omelianchuk-etal-2024-pillars} & 73.8 & 50.4 & \underline{67.5} & 60.5 & 48.6 & 57.7 & 75.2 & 70.0 & 74.1 \\
        
        \midrule
        \textbf{PoCO-\textit{Seq}} & 70.2 & 48.9 & 64.6 & \textbf{63.0} & 42.4 & 57.4 & \textbf{78.6} & 64.3 & \underline{75.3} \\
       \textbf{PoCO-\textit{Mix}} & 69.5 & {51.9} & 65.1  & \underline{62.3} & 45.3 &\underline{57.9}  & \underline{78.0} & {67.8} & \textbf{75.7} \\
        \bottomrule
    \end{tabular}
    }
    \caption{Performance comparison between LLM-based models and PoCO (based on the T5-large model) across different test sets. \textbf{Bold} and \underline{underlined} texts indicate the highest and the second highest value, respectively. }
    \label{tab:performance3}
\end{table*}

\begin{table*}[htbp]
    \centering
    \scalebox{0.73}{
    \begin{tabular}{l|ccc|l|ccc}
        \toprule
        \multirow{2}{*}{\textbf{No. of Base Systems}} & \multicolumn{3}{c|}{\textbf{ESC}} & \multirow{2}{*}{\textbf{No. of Base Systems}} & \multicolumn{3}{c}{\textbf{Ours (ensemble)}} \\ \cline{2-4} \cline{6-8}
        & \textbf{Precision} & \textbf{Recall} & \textbf{F$_{0.5}$} & & \textbf{Precision} & \textbf{Recall} & \textbf{F$_{0.5}$} \\ \midrule

        T5-Large + GECTor XLNet & 80.2 & 61.1 & 75.5 & PoCO-\textit{Mix} 
 + Triggered Overcorrection & 84.6 & 61.8 & 78.8 \\ 
        + GECToR Roberta & 84.7 & 59.0 & 77.9 & + PoCO-\textit{Gold} & 85.2 & 61.5 & 79.1 \\ 
        + Riken \& Tohoku & 86.2 & 59.4 & 79.0 & + GECTor XLNet & 85.8 & 63.2 & 80.0 \\ 
        + UEDIN-MS & 86.2 & 61.1 & 79.6 & + GECTor Roberta & 86.1 & 62.9 & 80.2 \\ 
        + Kakao \& Brain & 86.7 & 60.9 & \textbf{79.9} & + Riken \& Tohoku & 86.2 & 63.6 & \textbf{80.5} \\ \midrule
        & & & & + UEDIN-MS & 87.6 & 62.6 & 81.1 \\ 
        & & & & + Kakao \& Brain & 87.5 & 63.0 & \textbf{81.2} \\ \bottomrule

    \end{tabular}
    }
    \caption{Comparison of the original ESC and our ensemble results using ESC with different base system combinations. Models in each row are incrementally added. Our PoCO results (PoCO-\textit{Mix}, PoCO-\textit{Gold}) correspond to large-model settings.}
    \label{tab:our_esc_model}
\end{table*}

\paragraph{Effects of recovered targets}
In addition to the two methods proposed in Section 3.2, we also conduct experiments using only the gold target and the recovered target, respectively. 
The results for each training approach are presented in Table \ref{tab:results5}. 
Our findings clearly demonstrate the advantages of training with the recovered targets we proposed. 
When trained solely with gold targets, the model generally exhibits high recall, but when compared to using the recovered target for training, the precision tends to show low scores, ultimately leading to a decline in $F_{0.5}$ score.
In contrast, training exclusively with recovered targets leads to a substantial improvement in precision scores, significantly enhancing the reliability of the model. 
When the recovered target is combined with the gold target, it consistently achieves the best performance on $F_{0.5}$ scores across nearly all experiments. Notably, in the PoCO-\textit{Mix} setting, both precision and recall improved in a balanced manner. 
Additionally, there is a significant difference in recall between the base model and the large model, with the large model generally being more effective in preserving the high recall achieved by the LLM. 
For precision, the experimental results using the recovered target show minimal variation across different model sizes (T5-base vs. T5-large). 
This indicates that our method consistently contributes to improving precision regardless of model scale, demonstrating its robustness and effectiveness across different parameter sizes.

\paragraph{Reference-free evaluation}

We additionally report reference-free evaluation using GPT-4.1 as an automatic judge, following the framework of \citet{kobayashi-etal-2024-large} with minor modifications to the BEA-19 dev setup. GPT-4.1 directly assessed 50 randomly sampled system outputs on a 1–5 scale for \textit{grammaticality}, \textit{fluency}, and \textit{meaning preservation}, thereby capturing aspects often missed by overlap-based metrics \cite{grundkiewicz-etal-2015-human, sakaguchi-etal-2016-reassessing, yoshimura-etal-2020-reference, gong-etal-2022-revisiting, kobayashi-etal-2024-large, ostling-etal-2024-evaluation}. Both PoCo-\textit{Mix} and PoCo-\textit{Seq} achieved consistently high scores (over 4.7 in all dimensions), with both reaching 4.98 in meaning preservation, confirming that our method produces outputs of high linguistic quality and semantic fidelity. Details are shown in Appendix~\ref{appendix}.


\section{Discussions and Analysis}


\paragraph{Comparison with LLMs}
Table \ref{tab:performance3} demonstrates the effectiveness of our method compared to LLM-based few-shot and fine-tuning approaches. Providing few-shot prompting to LLMs can enhance recall, but it does not help to improve precision, which remains a critical limitation. 
However, our approach, which leverages supervised sLMs with 770M parameters, exhibits competitive performance even when compared to directly fine-tuned LLMs with over 7 billion parameters. 
Notably, our approach achieves the highest precision on both the BEA-19 test and dev sets, and even obtaining the highest $F_{0.5}$ score on the BEA-19 test set.




\paragraph{PoCO with ESC}
Recently, models that achieve strong performance in GEC on $F_{0.5}$ scores are primarily based on ensemble methods which combines multiple models.
We aimed to demonstrate the effectiveness of our model by applying it to a high-performing ESC system, which ensembles robust fine-tuned model by using a simple linear regression method.
To show a direct comparison with previously published results, we conducted experiments exclusively on the BEA-19 test set.
To ensure model diversity, we incorporated three models: LLM with an overcorrection prompt, PoCO-\textit{Gold}, and PoCO-\textit{Mix}.
As shown in Table \ref{tab:our_esc_model}, applying our proposed three models resulted in superior performance compared to the three models used in the existing ESC system.
Furthermore, when we incorporated the GECToR-XLNet model used in the ESC system into ours, we achieved a performance of 80.0 using only four models, surpassing the 79.9 score obtained with the original six-model ensemble.
When combining our models with those used in the ESC system, we achieved a performance of 81.2. This result is competitive with the 81.4 scores reported in a state-of-the-art study \cite{omelianchuk-etal-2024-pillars}, which uses fine-tuned LLMs for the ensemble.

\begin{figure}
\centering
\includegraphics[width=\linewidth]{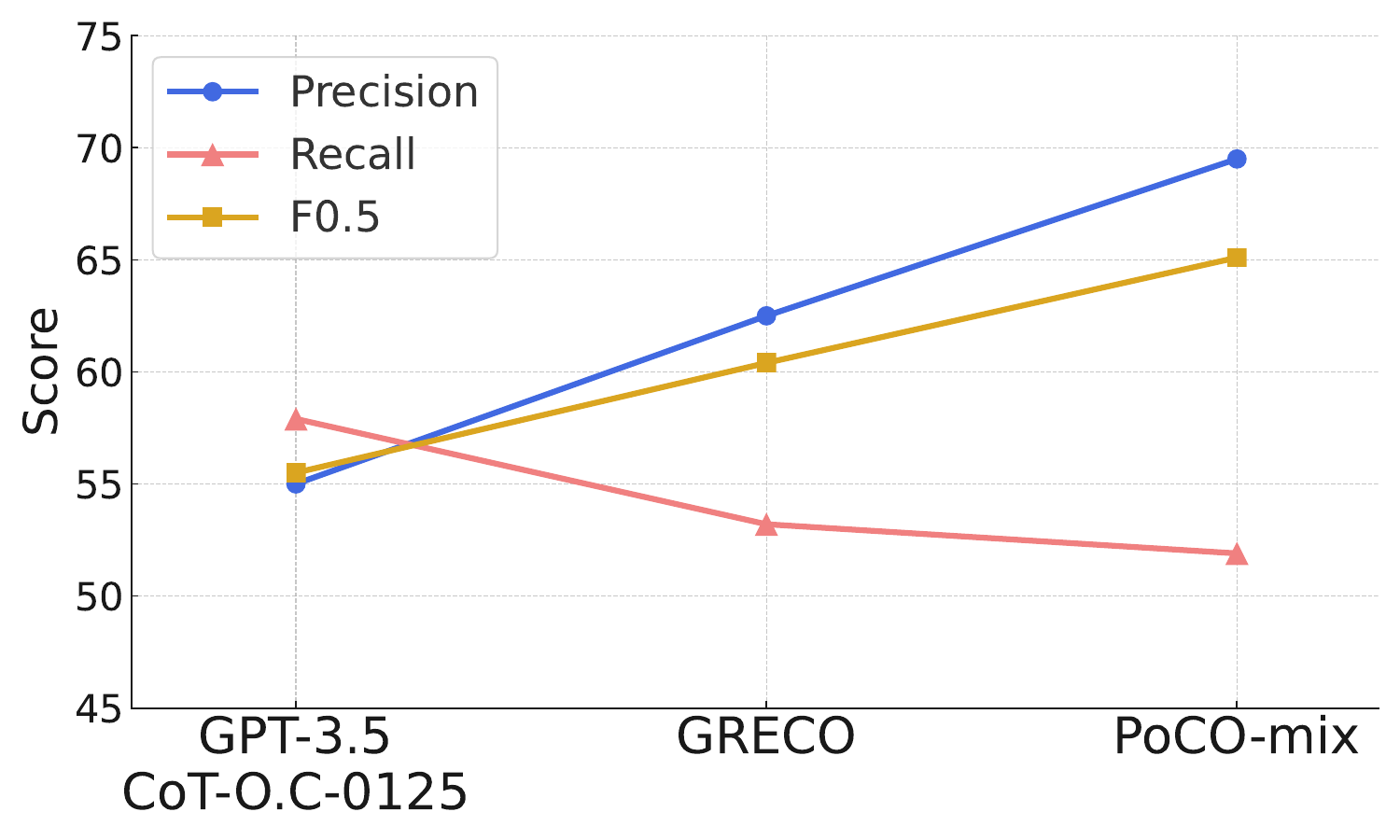}
\caption{Illustration of performance comparison with GRECO, showing that PoCO substantially outperforms in both precision and $F_{0.5}$ on the CoNLL-14 test set.}%
\label{fig:GRECO}
\end{figure}

\paragraph{PoCO vs GRECO}
To evaluate the effectiveness of PoCO against existing quality estimation methods, we conduct additional experiments using the top-5 GPT-generated corrections instead of the Riken\&Tohoku outputs. As shown in Figure~\ref{fig:GRECO}, PoCO-\textit{Mix} outperforms GRECO on the CoNLL-14 test set, achieving notably higher precision (+7.0) and $F_{0.5}$ (+4.9). Although its recall is slightly lower than that of GRECO, PoCO-\textit{Mix} still remarkably improves recall over its comparison sLMs (Table~\ref{tab:performance4}), demonstrating a better precision–recall trade-off and superior $F_{0.5}$ score overall.

\definecolor{yel}{RGB}{255, 240, 153}

\begin{table}[t]
    \centering
    \scalebox{0.70}{
    \begin{tabular}{ p{10cm} }
        \toprule
        \textbf{LLM Post Correct Prompt}\\
        \hline
        Please identify and correct any grammatical errors in the source sentence while avoiding unnecessary changes (overcorrections) and insufficient edits (undercorrections).

Correction Types:

- Source: The original sentence before any corrections.

- Overcorrect: Contains unnecessary modifications but also many correct edits, leading to high recall but low precision.

- Undercorrect: Makes fewer incorrect changes but misses valid corrections, resulting in higher precision but lower recall.

Hint Tags:

- <R>...</R>: Indicates a replaced part of the sentence.

- <M>...</M>: Marks an addition that wasn't in the source.

- <U>...</U>: Highlights a removed part of the original sentence.

Your task is to provide the best possible correction for the given source sentence, ensuring proper grammar and clarity while preserving its intended meaning. Use the "Overcorrect" (high recall) and "Undercorrect" (high precision) sentences, along with hint tags, to guide your edits.

Input Data: 

Source: \{\{Source\}\}

Overcorrect: \{\{Overcorrected output\}\}

Undercorrect: \{\{Undercorrected output\}\}

Perfect Correction: \\
        \bottomrule
    \end{tabular}
    }
    \caption{The prompt for Post-correction with LLMs.}
    \label{tab:prompting2}
\end{table}

\paragraph{LLMs for post-correction}

We investigated whether LLMs alone could effectively perform post-correction without relying on our proposed PoCO method. To evaluate this, we conducted experiments using GPT-4o-mini and GPT-4-Turbo, testing their ability to correct overcorrections made by LLMs. In particular, we offer information on overcorrection and undercorrection when providing a fine-tuned T5-base model's output to LLM, so that it can recognize what to recover.
Additionally, inspired by the findings of \cite{ryu24_interspeech}, which demonstrated that highlighting key areas improves model performance, we tagged crucial segments with the following tags: \texttt{<R>}, \texttt{<M>}, \texttt{<U>}, which indicate \textit{Replace, Missing, Unnecessary} words, respectively. Specifically, we marked modifications between the source sentence and the LLM-overcorrected output, as well as between the source sentence and the high-precision T5 output, giving the model explicit correction cues. The full prompt for this experiment is detailed in Table \ref{tab:prompting2}.

\begin{figure}
\centering
\includegraphics[width=\linewidth]{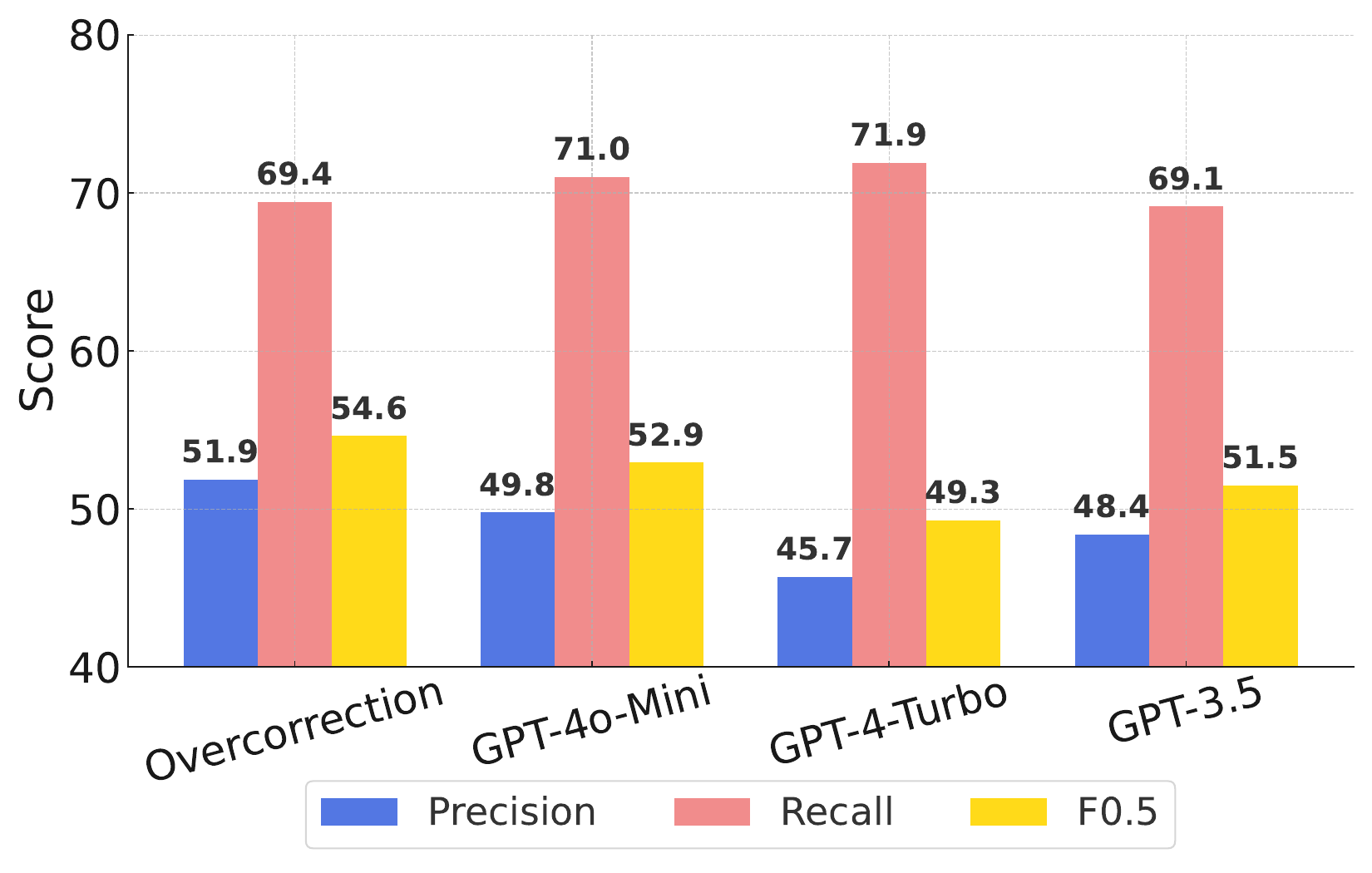}
\caption{Precision, recall, and $F_{0.5}$ scores on BEA-19 test set when using LLMs (\textit{GPT-4o-Mini, GPT-4-Turbo, and GPT-3.5}) for post-correction after our triggered \textit{Overcorrection}, instead of our PoCO method.}%
\label{fig:LLM_correction}
\end{figure}

\begin{figure}
\centering
\includegraphics[width=\linewidth]{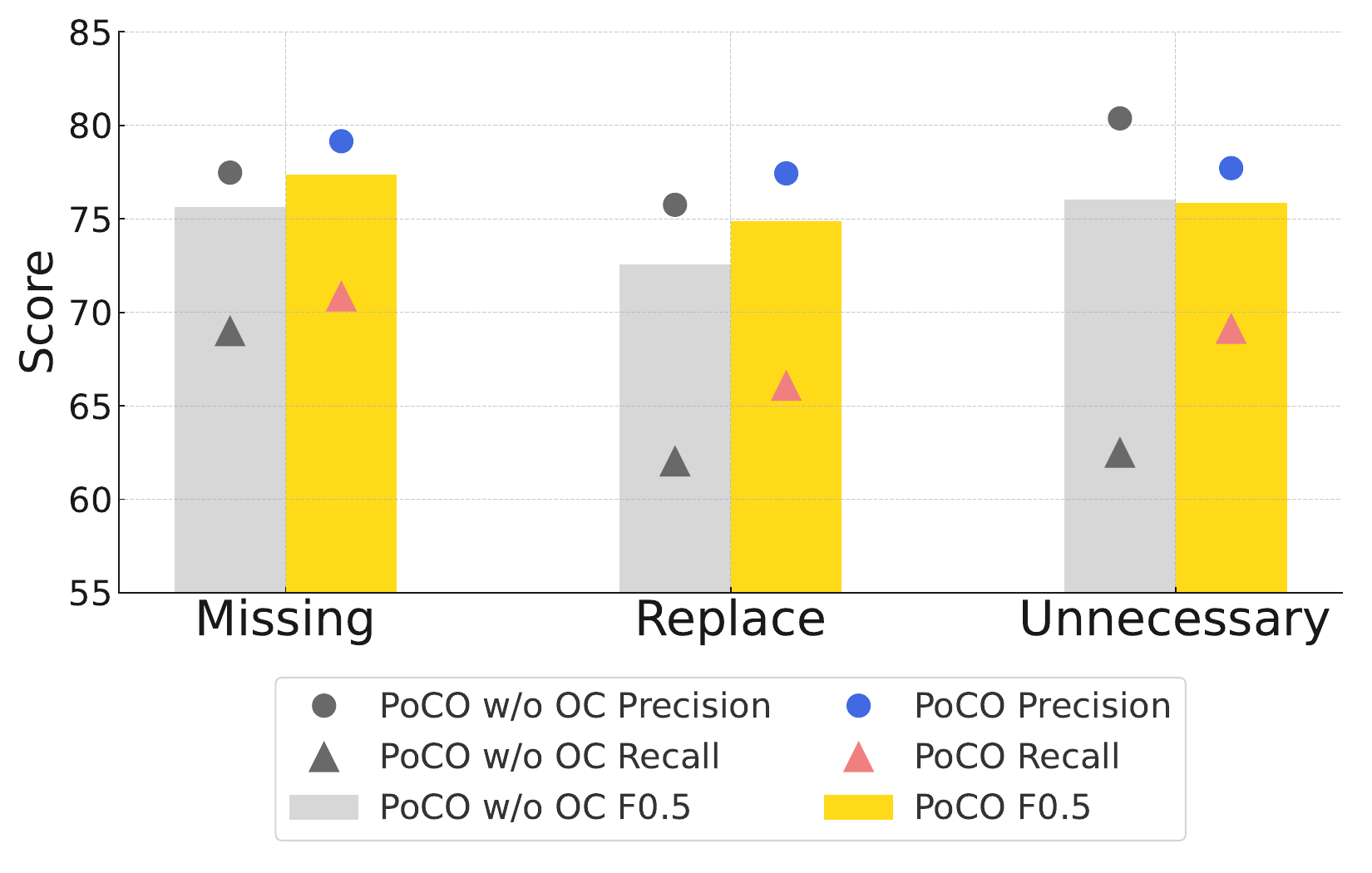}
\caption{Performance comparison of PoCO w/o overcorrection and PoCO by error type (\textit{Missing, Replace, Unnecessary}).}%
\label{fig:poco_comparison}
\end{figure}

We expected this setup to somewhat assist LLMs in balancing under-correction and overcorrection, producing a well-revised output; however, the results in Figure \ref{fig:LLM_correction} indicate different results. Instead of improving, LLM-based correction actually led to a drop in overall performance. While recall increased, precision declined, making the final output less reliable. Notably, GPT-4-Turbo suffered an even greater performance drop compared to GPT-4o-mini. These findings further support a key takeaway that LLMs alone struggle with overcorrection in GEC tasks, as previously reported in prior studies \cite{loem2023exploringeffectivenessgpt3grammatical, fang2023chatgpthighlyfluentgrammatical, wu2023chatgptgrammarlyevaluatingchatgpt, coyne2023analyzingperformancegpt35gpt4}.
PoCO, on the contrary, successfully tackles this issue, proving to be far more effective than simply relying on LLM-based correction.

\paragraph{Analysis of error types}
We further analyze results by error type, i.e., \textit{Missing}, \textit{Replace}, and \textit{Unnecessary} (Figure~\ref{fig:poco_comparison}). \textit{Missing} means essential words are omitted, \textit{Replace} refers to errors where an incorrect word or form must be substituted, and \textit{Unnecessary} indicates redundant words that should be deleted. These results are based on the official BEA-19 Shared Task evaluation, where system submissions automatically provide error-type–level performance.
We removed the intentional overcorrection step and compared the resulting model with the full PoCO pipeline on the BEA-19 test set.
Our overcorrection strategy consistently improves recall across all error types, fulfilling its intended goal. While \textit{Unnecessary} errors show a moderate trade-off with precision, the gains in recall remain substantial. Notably, in \textit{Missing} and \textit{Replace} cases, recall gains were accompanied by higher precision, showing that first guiding the model toward overcorrection is particularly effective at recovering omitted content and fixing incorrect tokens.

\section{Conclusion}

In this work, we introduced Post-Correction via Overcorrection (PoCO), a novel framework designed to address the precision-recall trade-off in grammatical error correction. By leveraging LLM-driven overcorrection to maximize recall and introducing targeted post-correction using fine-tuned smaller models to refine erroneous outputs, PoCO effectively balances precision and recall. Our extensive experiments demonstrate that PoCO enhances overall GEC performance by successfully mitigating overcorrection while maintaining the precision robustness of supervised models. These findings highlight the potential of integrating LLMs and smaller models for more accurate and reliable grammatical error correction. 


\section*{Limitations}

We have explored LLMs' ability to handle overcorrection; however, there is currently a lack of dedicated studies investigating how different prompting techniques influence overcorrection in LLMs.
Consequently, we adopted prompting methods from prior studies, which primarily focused on general GEC rather than explicitly addressing overcorrection.
While our approaches provide a reasonable baseline for guiding overcorrection, they may not be fully optimized.


Also, our findings indicate that LLMs alone face challenges in effectively correcting overcorrection errors in GEC tasks. This highlights the necessity of further research into prompting strategies that can more effectively guide LLMs.
Given the lack of systematic studies on how different prompting techniques influence LLMs' ability to mitigate overcorrection errors, we anticipate that exploring more targeted and adaptive prompting methods will enhance LLM performance in post-correction tasks, leading to improvements in both recall and precision in GEC outputs.

\section*{Acknowledgments}
This research was partly supported by the MSIT (Ministry of Science and ICT), Korea, under the ITRC (Information Technology Research Center) support program (IITP-2025-RS-2020-II201789) supervised by the IITP (Institute for Information \& Communications Technology Planning \& Evaluation; 47.5\%), by Culture, Sports and Tourism R\&D Program through the Korea Creative Content Agency grant funded by the Ministry of Culture, Sports and Tourism in 2025 (Project Name: Development of an AI-Based Korean Diagnostic System for Efficient Korean Speaking Learning by Foreigners, Project Number: RS-2025-02413038; 47.5\%), and by Institute of Information \& communications Technology Planning \& Evaluation (IITP) grant funded by the Korea government(MSIT) (No.RS-2019-II191906, Artificial Intelligence Graduate School Program (POSTECH); 5\%).

\section*{Ethical Statement}
This study exclusively utilizes publicly available datasets of grammatical error correction, where personal or other sensitive information is not included, adhering to ethical guidelines and policies. 


\bibliography{custom}
\appendix

\section{Reference-free Evaluation}\label{appendix}

The full prompt used for the reference-free evaluation described in Section~5.3 is provided in Table~\ref{tab:prompting3}. 
For the trait-specific instruction, we directly adopted the instruction prompts from \citet{kobayashi-etal-2024-large}. 
However, unlike their setup which presents multiple target sentences at once, 
our evaluation was conducted by providing a single source sentence and a single target sentence for scoring. 
The complete results are reported in Table~\ref{tab:reference_less_eval}.

\begin{table}[H]
    \centering
    \scalebox{0.70}{
    \begin{tabular}{ p{10cm} }
        \toprule
        \textbf{Modified Reference-free Evaluation Prompt}\\
        \hline
The goal of this task is to rank the presented targets based on the quality of the sentences. 
The source input consists of a single sentence written by an English learner. 
Please assign a score from 1 point to 5 points to each target based on the quality of the sentence (note that you can assign the same score multiple times). \\
\{trait-specific instruction\} \\

\# source \\
\{source input\} \\

\# target sentences \\
\{target sentence\} \\

\# output format \\
The output should be a markdown code snippet formatted in the following schema, including the leading and trailing "```json" and "'''": \\

```json
\{ \\
"target\_score": int // assigned score for target sentence \\
\} \\
'''
 \\
        \bottomrule
    \end{tabular}
    }
    \caption{Prompts used for reference-free evaluation. We adapt the original essay-level setting into a sentence-level prompt suitable for BEA-19 dev evaluation.}
    \label{tab:prompting3}
\end{table}

\begin{table}[H]
    \centering
    \scalebox{0.7}{
    \begin{tabular}{lccc}
        \toprule
        \textbf{Model} & \textbf{Fluency} & \textbf{Grammaticality} & \textbf{Meaning} \\
        \midrule
        PoCO-\textit{Mix}       & 4.70 & 4.78 & 4.92 \\
        PoCO-\textit{Seq}       & 4.72 & 4.72 & 4.92 \\
        PoCO-\textit{Recovered} & 4.74 & 4.80 & 4.92 \\
        PoCO-\textit{Gold}      & 4.68 & 4.80 & 4.96 \\
        \bottomrule
    \end{tabular}
    }
    \caption{Reference-free evaluation results (1–5 scale) across different PoCO-large variants.}
    \label{tab:reference_less_eval}
\end{table}



\end{document}